\documentclass{article}

\usepackage{microtype}
\usepackage{graphicx}
\usepackage{subfigure}
\usepackage{booktabs} 

\usepackage{hyperref}
\usepackage{colortbl}
\usepackage{amsmath}

\usepackage[accepted]{icml2024}

\usepackage{amsmath}
\usepackage{amssymb}
\usepackage{mathtools}
\usepackage{amsthm}

\usepackage[capitalize,noabbrev]{cleveref}

\theoremstyle{plain}

\theoremstyle{definition}

\theoremstyle{remark}

\usepackage[textsize=tiny]{todonotes}

\icmltitlerunning{Capacity Control is an Effective Memorization Mitigation Mechanism}

\DeclareMathOperator{\argmin}{arg\,min}

\begin{document}

\twocolumn[
\icmltitle{Capacity Control is an Effective Memorization Mitigation Mechanism in Text-Conditional Diffusion Models}

\icmlsetsymbol{equal}{*}

\begin{icmlauthorlist}
\icmlauthor{Raman Dutt}{yyy}
\icmlauthor{Pedro Sanchez}{xxx}
\icmlauthor{Ondrej Bohdal}{yyy}
\icmlauthor{Sotirios A. Tsaftaris }{xxx}
\icmlauthor{Timothy Hospedales}{yyy,comp}
\end{icmlauthorlist}

\icmlaffiliation{yyy}{School of Informatics, University of Edinburgh, Edinburgh, UK}
\icmlaffiliation{xxx}{School of Engineering, University of Edinburgh, Edinburgh, UK}
\icmlaffiliation{comp}{Samsung AI Center, Cambridge, UK}

\icmlcorrespondingauthor{Raman Dutt}{raman.dutt@ed.ac.uk}

\icmlkeywords{Diffusion Models, Memorization, PEFT, Medical Image Analysis}

\vskip 0.3in
]

\printAffiliationsAndNotice{\icmlEqualContribution} 



\section{Introduction and Related Work}
\label{submission}

Generative diffusion models \cite{sohl2015deep, song2021denoising, ho2020denoising, song2021scorebased} excel in generating high-quality data across various formats, including images \cite{dhariwal2021diffusion, karras2019style, ramesh2022hierarchical, rombach2022high}, audio \cite{kim2022guided, huang2023make}, and graphs \cite{xu2022geodiff, vignac2023digress}. These models power successful applications like \textit{Midjourney}, \textit{Stable Diffusion}, and \textit{DALL-E}. The strength of these models lies in their ability to produce high-quality, novel images that capture real-world complexities. A notable application is using synthetic data to share sensitive information without privacy risks, thus facilitating progress. However, this benefit is compromised if models inadvertently reproduce training data, known as \textit{memorization}, raising legal and ethical concerns \cite{butterick_2023}. This issue has prompted the development of mitigation methods, including training-time interventions  \cite{webster2023duplication, somepalli2023understanding} noise regularizers \cite{somepalli2023diffusion}, and inference-time solutions \cite{somepalli2023understanding, wen2024detecting}.

In domain-specific diffusion models for medical imaging, the process typically involves starting with a general-purpose model and fine-tuning it with specific data \cite{chambon2022roentgen}. Our paper hypothesizes that the overcapacity of large neural networks leads to memorization. We explore regularizing model capacity during fine-tuning to mitigate this issue, using Parameter-Efficient Fine-Tuning (PEFT) methods from text-conditional diffusion models. Our results show that capacity-controlled approaches, like PEFT, effectively mitigate memorization during training in diffusion models.

\section{Capacity Control using Parameter-Efficient Fine-Tuning}

The conventional way of adapting a \emph{pre-trained} model to a specific domain is by updating the entire parameter set using a downstream dataset $\mathcal{D}^{train}$ and a task-specific loss $\mathcal{L}^{base}$. An alternate is offered by Parameter-Efficient Fine-Tuning (PEFT) methods that involve freezing most of the original \emph{pre-trained} model and fine-tuning a specific parameter subset. PEFT methods can be broadly categorized into \textit{Selective} and \textit{Additive} strategies. The \textbf{Selective} category includes methods that involve fine-tuning a subset of existing model parameters such as attention matrices \cite{touvron2022three}, normalization layers \cite{basu2023strong} and bias parameters \cite{bitfit}. The \textbf{Additive} category of methods freezes the original model and introduces a new set of parameters for fine-tuning \cite{lian2022scaling,chen2022adaptformer,han2023svdiff,hu2022lora,xie2023difffit}. For text-to-image generation, strategies such as SV-DIFF \cite{han2023svdiff} and DiffFit \cite{xie2023difffit} have been proposed, and they excel in both in-domain and out-of-domain settings \cite{dutt2023parameter}. 

The key idea of PEFT is to fine-tune a very small subset of model parameters $\phi \subset \theta$, i.e. $|\phi| \ll |\theta|$. The parameters of the pre-trained model are initially $\theta_0$, and after fine-tuning on data $\mathcal{D}^{train}$ they change by $\Delta\phi$ compared to their initial values. The fine-tuning process can then be defined via:


\begin{equation}
    \Delta\phi^* = \argmin_{\Delta\phi} \mathcal{L}^{base} \left( \mathcal{D}^{train}; \theta_0 + \omega \odot \Delta\phi  \right)
\end{equation}

where $\mathcal{L}^{base}$ is the mean squared error in the context of image generation problems.

\section{Experiments}

\subsection{Experimental Setup}
\textbf{Architecture:} Our experiments adopted the pre-trained Stable Diffusion (v1.5) \cite{Rombach_2022_CVPR} model implemented in the \textit{Diffusers} package \cite{von-platen-etal-2022-diffusers}. For fine-tuning on medical data we follow previous practice \cite{dutt2023parameter, chambon2022roentgen}, and fine-tune only the U-Net while keeping the other components in the stable diffusion pipeline frozen during our experiments. 

\textbf{Datasets: } We use the MIMIC dataset \cite{johnson2016mimic} that consists of chest X-rays and associated radiology text reports. The dataset was split into train, validation and test splits ensuring no leakage of subjects, resulting in 110K training samples, 14K validation and 14K test samples. Note that several images in this dataset can be associated with the same text prompt due to similarity in clinical findings.


\subsection{Memorization and Image-Quality Metrics} \label{sec:metrics}

\textbf{Memorization: Nearest Neighbour Distance}\quad A generated example $\hat{x}$ is considered memorized if it is very close to a training example $x$, measured by a distance metric $l$ with threshold $\delta$. We report the average minimum cosine distance between each synthetic sample and its nearest training example.
\textbf{Memorization: Extraction Attack}\quad We assess memorization using an image extraction attack \cite{carlini2023extracting}, which involves generating many images for a given prompt and performing membership inference to identify memorized images. Fewer extracted images indicate less memorization.
\textbf{Memorization: Denoising Strength Heuristic}\quad As a proxy for memorization, we use the magnitude of the diffusion noise prediction \cite{wen2024detecting}. This measures the difference between model predictions conditioned on a prompt embedding $e_{p}$ and an empty string embedding $e_{\phi}$ averaged over multiple time steps.
\textbf{Quality: Fréchet Inception Distance (FID)}\quad FID score evaluates image generation quality by comparing distributions of real and synthetic images based on features from a pre-trained image encoder. We use a DenseNet-121 model \cite{huang2017densely} pre-trained on chest X-rays \cite{torchxrayvision} for feature extraction.
\textbf{Quality: BioViL-T Score}\quad BioViL-T \cite{bannur2023learning}, a vision-language model for chest X-rays and radiology reports, provides a joint score indicating the clinical correlation between an image and a text caption. Higher scores reflect better quality and textual guidance of generated images.

\section{Results}

In this section, we report our results from the experiments performed on the MIMIC dataset for generation quality (FID $\downarrow$, BioViL-T Score $\uparrow$) and memorization (AMD $\uparrow$, Extracted Images $\downarrow$).

\textbf{Table \ref{tab:all_results_combined}} reports the results for comparing PEFT strategies with conventional fine-tuning methods. PEFT methods act as strong regularizers and hence, significantly restrict the model capacity during fine-tuning. The results indicate that controlling the model capacity can enable both high-generation quality and memorization mitigation. 

\textbf{Table \ref{tab:table2}} indicates results for combining different fine-tuning methods (conventional and PEFT) with existing memorization mitigation strategies \cite{somepalli2023understanding,wen2024detecting}. The results indicate that while existing mitigation strategies do play their part, they can easily be combined with PEFT methods to reduce memorization further and significantly outperform conventional fine-tuning.

\begin{table}[ht]
\centering
\caption{Comparitive evaluation of generation quality (FID $\downarrow$, BioViL-T Score $\uparrow$) and memorization (AMD $\uparrow$, Extracted Images $\downarrow$) on the MIMIC dataset. PEFT strategies consistently outperform conventional fine-tuning methods in terms of both generation quality and memorization.}
\resizebox{0.5\textwidth}{!}{%
\begin{tabular}{@{}lcccc@{}}
\toprule
\textbf{FT Method}                  & \textbf{FID Score ($\downarrow$)} & \textbf{AMD ($\uparrow$)} & \textbf{BioViL-T Score ($\uparrow$)} & \textbf{Extracted Images ($\downarrow$)} \\ \midrule
Freeze                             & 323.78                            & \textbf{0.801}                        & 0.23             & \phantom{00}0                    \\
Full FT                            & \phantom{0}35.79                             & 0.029                       & 0.60             & 356                    \\


DiffFit                          & \phantom{0}15.19                             & 0.061                        & 0.69       & \phantom{0}69                          \\
Attn Tuning          & \phantom{0}40.28                             & 0.031                       & 0.59       & 358                          \\
SV-DIFF   & \phantom{0}\textbf{12.77}                             & 0.062                        & \textbf{0.72}       & \phantom{0}\textbf{63}                        \\

 \bottomrule
\end{tabular}
}
\label{tab:all_results_combined}
\end{table}

\begin{table}[ht]
\centering
\caption{Combining different fine-tuning strategies combined with existing memorization mitigation mechanisms.}
\resizebox{0.5\textwidth}{!}{%
\begin{tabular}{@{}lcccc@{}}
\toprule
\textbf{FT Method}                  & \textbf{FID Score ($\downarrow$)} & \textbf{AMD ($\uparrow$)} & \textbf{BioViL-T Score ($\uparrow$)} & \textbf{Extracted Images ($\downarrow$)} \\ \midrule

Full FT  + RWA                      & \phantom{0}67.48                             & 0.041                       & 0.58       & 320                          \\
Full FT  + Threshold Mitigation     & \phantom{0}25.43                             & 0.048                        & 0.62       & 298                           \\ \\

DiffFit  RWA                     & \phantom{0}17.18                             & 0.060                        & 0.67       & \phantom{0}64                          \\
DiffFit  Threshold Mitigation     & \phantom{0}\textbf{15.31}                             & 0.067                        & \textbf{0.69}       & \phantom{0}51                          \\ \\

Attn Tuning  + RWA                 & \phantom{0}49.71                             & 0.035                       & 0.52       & 324                          \\
Attn Tuning  + Threshold Mitigation   & \phantom{0}25.43                             & 0.041                        & 0.62       & 299                          \\ \\
SV-DIFF  + RWA                     & \phantom{0}41.25                             & 0.059                        & 0.52       & \phantom{0}58                          \\
SV-DIFF  + Threshold Mitigation      & \phantom{0}23.64                             & \textbf{0.068}                       & 0.64       & \phantom{0}\textbf{45}                          \\

 \bottomrule
\end{tabular}
}
\label{tab:table2}
\end{table}



\section{Conclusion and Future Work}

In this work, we addressed memorization in diffusion models by controlling the model's capacity. Our experiments showed that model capacity significantly influences memorization. By employing parameter-efficient fine-tuning (PEFT), we were able to reduce memorization while maintaining high-generation quality. Additionally, we demonstrated that these fine-tuning methods can complement existing memorization mitigation strategies.

For future work, we aim to develop an automated scheme for selecting the optimal PEFT parameters during fine-tuning to balance memorization and generalization. Similar frameworks have been proposed for optimizing parameter updates based on reward metrics \cite{dutt2024fairtune}. We plan to formulate our approach as a bi-level optimization problem, where the outer loop proposes parameter subsets to update, and the inner loop fine-tunes these parameters on the downstream task. The process will be guided by reward metrics that assess both generation quality and memorization. In this process, our framework can find the optimal model capacity (parameter subsets) to update to improve both generation quality and reduce memorization.

\bibliography{example_paper}
\bibliographystyle{icml2024}

\newpage
\appendix
\onecolumn

\end{document}